%% file: main.tex
\documentclass[runningheads]{llncs}
\usepackage[T1]{fontenc}
\usepackage{graphicx,verbatim}
\usepackage{hyperref}
\usepackage{amsmath} 
\usepackage{amssymb} 
\usepackage{cleveref}
\usepackage{booktabs}
\usepackage{multirow}
\usepackage{mathtools}

\begin{document}
%
\title{GRASPing Anatomy to Improve Pathology Segmentation}

\author{
Keyi Li\inst{1*} \and
Alexander Jaus\inst{1,2*,\dagger} \and
Jens Kleesiek\inst{3,4} \and
Rainer Stiefelhagen\inst{1}
}  
\authorrunning{Anonymized Author et al.}
\institute{
Karlsruhe Institute of Technology, Karlsruhe, Germany \and
Helmholtz Information and Data Science School for Health, Germany \and
Department of Nuclear Medicine, University of Duisburg-Essen \and
German Cancer Consortium (DKTK)- University Hospital Essen, Essen, Germany 
\email{alexander.jaus@kit.edu} 
}

\maketitle              
\footnotetext[1]{These authors contributed equally to this work. $\dagger$ Corresponding}
\begin{abstract}
Radiologists rely on anatomical understanding to accurately delineate pathologies, yet most current deep learning approaches use pure pattern recognition and ignore the anatomical context in which pathologies develop. To narrow this gap, we introduce GRASP (Guided Representation Alignment for the Segmentation of Pathologies), a modular plug-and-play framework that enhances pathology segmentation models by leveraging existing anatomy segmentation models through pseudolabel integration and feature alignment. Unlike previous approaches that obtain anatomical knowledge via auxiliary training, GRASP integrates into standard pathology optimization regimes without retraining anatomical components. We evaluate GRASP on two PET/CT datasets, conduct systematic ablation studies, and investigate the framework's inner workings. We find that GRASP consistently achieves top rankings across multiple evaluation metrics and diverse architectures. The framework's dual anatomy injection strategy, combining anatomical pseudo-labels as input channels with transformer-guided anatomical feature fusion, effectively incorporates anatomical context. The code is available at~\href{https://github.com/unrpi18/GRASP}{https://github.com/unrpi18/GRASP}.
\keywords{Lesion Segmentation  \and Anatomical Guidance \and PET/CT }

\end{abstract}
\input{Latex_source/Intorduction}
\input{Latex_source/Methodology}
\input{Latex_source/Evaluation}
\input{Latex_source/Conclusion}

\bibliographystyle{splncs04}
\bibliography{bibliography}

\end{document}

%% file: Latex_source/Intorduction.tex
\section{Introduction}
\label{se:Introduction}
When evaluating radiological scans for pathological findings, physicians heavily rely on an in-depth understanding of the human anatomy. This understanding enables them to more accurately distinguish between expected anatomical structures and unexpected findings that could potentially indicate pathological causes. The anatomically driven reasoning described stands in stark contrast to the pattern-recognition-based decisions of encoder-decoder style networks~\cite{cciccek20163d,oktayattention}, which typically rely on supervised training based on pathology-specific datasets and are often optimized in a single-task setting~\cite {gatidis2023autopet,oreiller2022head,menze2014multimodal}. While these methods have arguably achieved good performances~\cite{isensee2021nnu}, they often ignore the context in which pathology detection happens: the human anatomy, thereby ignoring the interrelatedness of anatomy and pathologies.
Recently, the field has made impressive progress in holistic anatomical segmentation~\cite{he2025vista3d,jaus2024towards,wasserthal2023totalsegmentator} with networks being able to capture large parts of the anatomy, yet the fields of anatomy and pathology segmentation largely remain distinct. In this work, we ask the question:
Can we leverage recent advancements in anatomical segmentation models and their embedded anatomical knowledge to enhance pathology segmentation models, thereby aligning them more closely with human expert workflows?

Within this work, we explore different strategies to align anatomical knowledge with the training of pathology segmentation models. Our contributions are: 1) We develop GRASP (Guided Representation Alignment for the Segmentation of Pathologies), a plug-and-play framework that leverages existing anatomy segmentation models to enhance pathology segmentation via pseudo-label integration and anatomical feature alignment. 2) We extensively test GRASP on two challenging PET/CT datasets across multiple splits using four distinct metrics. 3) We analyze GRASP through ablation studies, examining feature similarities and quantifying the influence of different anatomical models on segmentation performance. 4) We discuss related strategies for incorporating anatomical knowledge into pathology segmentation models and address challenges.

\section{Related Work}

While anatomical knowledge has been extensively explored for anatomy segmentation, its application to pathology segmentation remains less investigated. Early approaches incorporated specialized anatomical priors for specific pathologies: location priors for pancreatic cancer segmentation~\cite{dong2025position}, region-guided approaches for chest X-ray pathology detection~\cite{muller2023anatomy}, and anatomy-guided patchification for colorectal cancer segmentation~\cite{zhang2023ag}. These methods showed domain-specific promise but remained highly specialized and difficult to generalize.

Recent work has shifted toward incorporating anatomical pseudo-labels directly into model training. APEx~\cite{jaus2024anatomy} introduced dual-decoder joint segmentation for PET-CT and X-ray modalities, but operates in 2D thus ignoring crucial volumetric interrelations. Multi-label approaches~\cite{murugesan2023improving} gained traction in AutoPET challenges, simultaneously predicting anatomical and tumor classes, but require careful organ selection and loss weighting to artificially focus on pathology classes~\cite{kalisch2024autopet} since standard losses lack inherent class preferences. Extensive multi-modal pre-training with dual-decoder architectures~\cite{rokuss2024fdg} showed improvements but demands complex dataset curation, loss balancing, and large computational requirements. These methods share common limitations: They require fundamental pipeline modifications or necessitate auxiliary training losses and extensive pre-training to learn anatomical representations from scratch. This raises a fundamental question: why reinvent anatomy segmentation as an auxiliary task when highly capable anatomical models already exist? 

%% file: Latex_source/Methodology.tex
\section{Methodology}
\label{se:Methodology}
We review a series of previously mentioned strategies to incorporate anatomical knowledge into the pathology segmentation model training. We discuss their challenges before introducing the proposed GRASP framework. For these preliminary experiments, we use the 3D-UNet~\cite{cciccek20163d} architecture due to its popularity and experiment on the AutoPET~\cite{gatidis2023autopet} dataset, with PET/CT input $x \coloneqq (x^{\text{ct}}, x^{\text{pet}})$.
\subsection{Exploratory Strategies for Anatomy-Pathology Alignment}

\noindent \textbf{Transfer via Fine-Tuning:} Fine-tuning naturally arises as a first approach when aiming to leverage anatomical knowledge for pathology segmentation. It offers a simple way to reuse spatial and semantic representations from well-pretrained anatomy models. Specifically, we fine-tune a pretrained anatomy model $f_{\theta_{\text{ana}}}$, originally trained on multi-organ labels $y^{\text{ana}} \in \{0,1,\dots,\mathcal{C}_{\text{ana}}\}$ from the \textit{DAP Atlas}~\cite{jaus2024towards}, with $\mathcal{C}_{\text{ana}} \gg 2$. These rich anatomical features are expected to benefit the target binary task $(y^{\text{path}} \in \{0, 1\})$ despite the domain gap. We initialize $\theta \leftarrow \theta_{\text{ana}}$ and fine-tune on the pathology dataset $\mathcal{D}$:
\begin{equation}
\label{eq:finetuning}
\theta_{\text{path}}^* = \arg\min_{\theta} \sum_{(x_i, y_i^{\text{path}}) \in \mathcal{D}} \mathcal{L}(f_\theta(x_i), y_i^{\text{path}}).
\end{equation}
The pretrained anatomy model is typically trained with CT-only input~\cite{wasserthal2023totalsegmentator,jaus2024towards}. Accordingly, we retain the pretrained weights for the CT channel and randomly initialize the PET channel. We consider two fine-tuning configurations: 1) a full 300-epoch schedule using the same learning rate as the baseline, and 2) a shorter 100-epoch schedule with a reduced learning rate. We report the results in~\Cref{tab:finetuning} and observe that both approaches lead to a degraded performance compared to training from scratch. While initially surprising, this confirms similar results~\cite{kalisch2024autopet} and likely reflects the substantial class distribution shift. This underscores a key limitation: meaningful transfer requires careful and diverse dataset curation~\cite{rokuss2024fdg}, which complicates the development of simple, generalizable solutions.

\noindent \textbf{Multi-Class Supervision Strategy:} While fine-tuning enables the reuse of anatomical representations, it maintains a strict separation between anatomy and pathology supervision. To more directly incorporate anatomical knowledge into the learning process, we adopt a unified multi-class formulation with a shared label space. Specifically, we define 
\(
y^{\text{multiclass}} \in \{0,1,\dots,\mathcal{C}_{\text{ana}}, c_{\text{path}}\},
\)
where the tumor class \(c_{\text{path}}\) is appended as the last index. Anatomical pseudo-labels are derived from \textit{TotalSegmentator}~\cite{wasserthal2023totalsegmentator}, with fine-grained substructures (e.g., individual lung segments) consolidated into one label per anatomical region to reduce GPU memory consumption. 
We compare three loss weighting strategies: \textit{Standard} (uniform), \textit{Tumor-Focused} (tumor upweighted), and \textit{Patch-Aware}. The \textit{Patch-Aware} loss dynamically adjusts class weights per patch: absent classes are downweighted, anatomical classes receive moderate weight, and the tumor class is assigned the highest weight, promoting effective learning from both anatomy and pathology. Training follows an optimization objective similar to~\Cref{eq:finetuning}.
We observe that under the standard loss, the results degrade compared to the baseline model. Only when we modify the loss to significantly enhance the importance of the tumor class under the \textit{Patch-Aware} Loss setting does this approach slightly outperform the baseline; however, this marginal gain comes at the cost of substantial manual tuning and increased training complexity.

\noindent \textbf{Multi-Task Learning Approach:} To avoid coupling anatomical and pathological supervision in the multi-class formulation, we investigate a dual-branch architecture with a shared encoder $E_{\theta_{\text{enc}}}$ and two task-specific decoders $D_{\theta_{\text{ana}}}$ and $D_{\theta_{\text{path}}}$ for anatomy and pathology, respectively. This design enables task-specific predictions by decoding shared encoder features into separate anatomical and pathological outputs, supporting multi-label predictions and thus organ-pathology composition at inference time. The model is trained to simultaneously predict anatomical structures and tumor regions using separate output heads. The ground truth consists of two label maps: $y^{\text{path}} \in \{0, 1\}$ for binary tumor segmentation, and $y^{\text{ana}} \in \{0, 1, \dots, \mathcal{C}_{\text{ana}}\}$ for anatomical segmentation, where $\mathcal{C}_{\text{ana}}$ is the number of anatomical classes. The corresponding predictions are $\hat{y}_i^{\text{ana}} = D_{\theta_{\text{ana}}}(E_{\theta_{\text{enc}}}(x_i))$ and $\hat{y}_i^{\text{path}} = D_{\theta_{\text{path}}}(E_{\theta_{\text{enc}}}(x_i))$.
Training is guided by a joint loss function that balances the two tasks:
\[
\mathcal{L}_{\text{task}} = \alpha \cdot \mathcal{L}_{\text{path}} + (1 - \alpha) \cdot \mathcal{L}_{\text{ana}},
\]
where $\mathcal{L}_{\text{path}}$ is a binary segmentation loss for pathology, $\mathcal{L}_{\text{ana}}$ is a multi-class segmentation loss over merged anatomical pseudo-labels, and $\alpha \in [0, 1]$ controls the relative importance of pathology supervision.
To ensure a meaningful linear combination via $\alpha$, we normalize both loss terms by their means to match their magnitudes. Results and an $\alpha$ ablation are shown in~\Cref{tab:finetuning}. We find that this approach does outperform the baseline and previous approaches by a small margin. Interestingly, we observe that when the pathology loss weight is set too high ($\alpha = 0.95$), the optimizer tends to disregard the anatomical component, leading to a decline in performance for the pathology as well.

\input{exp/fine-tuning_results}

Overall, we find, that some of the explored approaches deliver minor improvements over the baseline model but require substantial parameter tuning. A key insight we find, is that all of these methods enforce auxiliary losses upon anatomical labels, but do have pathology segmentation as the primary goal, requiring a parameter-based importance increase for the pathology task. We thus raise the question: Can we remove the auxiliary anatomical training from the training process by leveraging well-performing anatomy segmentation models and thereby keeping the focus of the target model on pathology segmentation? 
In the following, we propose the GRASP framework as a solution.

\subsection{The GRASP Framework}
GRASP builds on the insight that high-quality anatomical models already exist. We introduce a plug-and-play framework that injects anatomical knowledge during pathology training by leveraging frozen anatomy encoders through feature alignment, without relying on auxiliary anatomical training.  In GRASP, the CT modality effectively serves a dual purpose: it is the input to the anatomy and the pathology model. We illustrate the overall framework in~\Cref{fig:GRASP_Arch}.

\input{Latex_source/Latex_figure/fig_grasp_arch}

\noindent \textbf{Dual Injection of Anatomical Priors:} Anatomical knowledge is integrated through two complementary mechanisms: 1) anatomical pseudo-labels as an auxiliary input channel, and 2) bottleneck-level feature fusion.
We follow a label-as-input strategy~\cite{jaus2024anatomy} by introducing a third input channel $x^{\text{ana}} \in \mathbb{R}^{H \times W \times D}$, which encodes voxel-wise anatomical pseudo-labels. We use $x^{\text{ana}}$ from a third model (e.g., \textit{TotalSegmentator}), though outputs from the frozen anatomy model are also viable.
For deeper integration, we also align and fuse anatomical features at the bottleneck level. Let the frozen pretrained anatomy model process the CT input to produce bottleneck features $\mathcal{Z}_{\text{ana}} = E_{\theta_{\text{ana}}}(x^{\text{ct}})$, while the pathology encoder extracts joint features from CT, PET, and anatomical pseudo-labels ($x^{\text{ct}}$, $x^{\text{pet}}$, $x^{\text{ana}}$), yielding $\mathcal{Z}_{\text{path}}$. As the spatial shape of $ \mathcal{Z}_{\text{ana}} $ and $ \mathcal{Z}_{\text{path}} $ may differ, we apply an \textit{Align Block} to adjust anatomical features. It consists of a 
pointwise convolution followed by adaptive spatial pooling, transforming $\mathcal{Z}_{\text{ana}} $ to match the dimensionality of $ \mathcal{Z}_{\text{path}} $. We then feed both $\mathcal{Z}_{\text{path}}$ and the aligned $\mathcal{Z}_{\text{ana}}$ into our proposed fusion module to perform feature-level integration guided by pathological features.

\noindent \textbf{Anatomy-Guided Transformer Fusion:} 
We design a lightweight fusion module leveraging transformer attention~\cite{vaswani2017attention}, as illustrated in~\Cref{fig:GRASP_Arch}. We first apply a spatial attention (SA) block inspired by CBAM~\cite{woo2018cbam} to emphasize informative spatial regions. For each voxel location $(h, w, d)$, attention weights are computed by aggregating feature responses across the channel dimension $C$ via average and max pooling, followed by a convolution and sigmoid activation. Next, we apply a channel-wise Squeeze-and-Excitation (SE)~\cite{hu2018squeeze} block to recalibrate the features, enhancing informative channels while suppressing less relevant ones. These two attention mechanisms are applied to both $\mathcal{Z}_{\text{ana}}$ and $\mathcal{Z}_{\text{path}}$, yielding recalibrated features $\hat{\mathcal{Z}}_{\text{ana}} \in \mathbb{R}^{B \times C \times H \times W \times D}$ and $\hat{\mathcal{Z}}_{\text{path}} \in \mathbb{R}^{B \times C \times H \times W \times D}$, where $B$ denotes the batch size:
\[
\hat{\mathcal{Z}}_{\text{ana}} = \text{SE}(\text{SA}(\mathcal{Z}_{\text{ana}})),
\hat{\mathcal{Z}}_{\text{path}} = \text{SE}(\text{SA}(\mathcal{Z}_{\text{path}})) ,
\]
where $\text{SA}(\cdot)$ and $\text{SE}(\cdot)$ denote the spatial and channel attention modules, respectively. The recalibrated pathological features $\hat{\mathcal{Z}}_{\text{path}}$ and anatomical features $\hat{\mathcal{Z}}_{\text{ana}}$ are reshaped into sequences $Q= \text{reshape}(\hat{\mathcal{Z}}_{\text{path}})$ and $K,V = \text{reshape}(\hat{\mathcal{Z}}_{\text{ana}})$, all with shape $\mathbb{R}^{B \times (H \cdot W \cdot D) \times C}$. We first apply self-attention to $Q$ to model intra-pathology dependencies, followed by cross-attention using $Q$ as queries and $K, V$ as keys and values. 
We adopt multi-head attention with $h=8$ heads~\cite{vaswani2017attention} to capture diverse anatomical-pathological relationships.
\noindent The result is reshaped back and fused with the original $\mathcal{Z}_{\text{path}}$ via a learnable gated sum:
$$
\tilde{\mathcal{Z}} = \delta \cdot O(Q, K, V) + (1 - \delta) \cdot \mathcal{Z}_{\text{path}}, \quad \delta = \texttt{Sigmoid}(w),
$$
where $O(Q, K, V)$ denotes the output of the \textit{Transformer Block}, with $\texttt{Sigmoid}(\cdot)$ controlling the contribution of the injected anatomical context, initialized to 0.5.

\noindent \textbf{Fusion Setup and Training Strategy:}
We select up to the two deepest convolutional layers from the pathology backbone encoder as the bottleneck block for feature fusion, fusing each with the corresponding layer from the anatomy encoder in a pair-wise manner. To simplify the design, we propose two strategies: a \textit{Mirror} setup, where a pathology model is trained from scratch and paired with a pretrained anatomy model of the same (mirrored) architecture trained on the \textit{DAP Atlas}~\cite{jaus2024towards}; and a \textit{Mixture} setup, where the same-architecture anatomy model is replaced with the off-the-shelf pretrained SegResNet~\cite{myronenko20183d} implemented by MONAI~\cite{cardoso2022monai}, improving generalizability. In practice, we employ a two-phase training strategy, where the fusion module is activated after 50 epochs, which has been shown to improve training stability.

%% file: exp/fine-tuning_results.tex
\begin{table}[ht]

\caption{Evaluation of pathology segmentation performances on AutoPET with a 3D-UNet backbone. LR indicates the learning rate; Loss FN, the loss function.
}
\centering
\scriptsize
\begin{tabular}{@{}lcccc@{}}
\hline
\textbf{Models} & 
\textbf{Epochs} & 
\textbf{LR} &
\textbf{Loss FN} &
\textbf{Dice~($\uparrow$)} \\
\hline

3D-UNet (baseline) & 300 & $1\text{e}{-4}$  & Standard DiceCE & 49.3 \\
\hline
Fine-tuning based 
       & 100 & $1\text{e}{-5}$  & Standard DiceCE & 36.6 \\
       & 300 & $1\text{e}{-4}$  & Standard DiceCE & 22.3 \\
\hline
Multi-class based  
           & 300 & $1\text{e}{-4}$ & Standard DiceCE & 45.0\\
           & 300 & $1\text{e}{-4}$ & Tumor-Focused DiceCE & 46.3 \\
           & 300 & $1\text{e}{-4}$ & Patch-Aware DiceCE & 49.6 \\
           \hline
Multi-task based 
           & 300 & $1\text{e}{-4}$ & normalized DiceCE ($\alpha=0.7$) & 51.0\\
           & 300 & $1\text{e}{-4}$ & normalized DiceCE ($\alpha=0.8$) & 51.8  \\
           & 300 & $1\text{e}{-4}$ & normalized DiceCE ($\alpha=0.9$) & 51.7 \\
           & 300 & $1\text{e}{-4}$ & normalized DiceCE ($\alpha=0.95$) & 47.6 \\
           \hline

\end{tabular}
\label{tab:finetuning}
\end{table}

%% file: Latex_source/Latex_figure/fig_grasp_arch.tex
\begin{figure}
\centering
\includegraphics[width=1.0\textwidth]{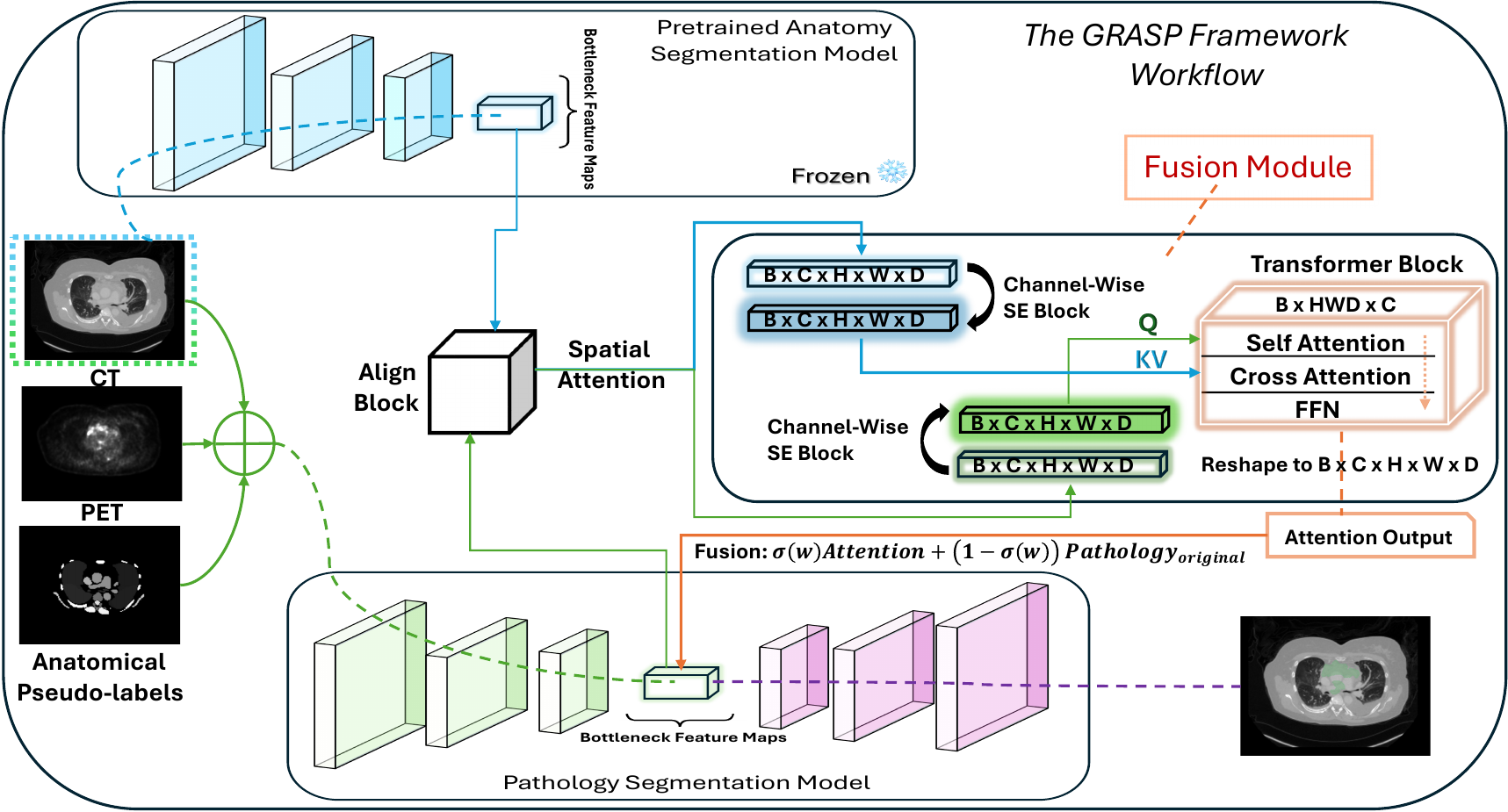}

\caption{GRASP framework: Bottleneck features from a frozen pretrained anatomy model are aligned and fused with pathology features via a modular fusion block.}
\label{fig:GRASP_Arch}
\end{figure}

%% file: Latex_source/Evaluation.tex
\section{Experiments and Results}
\label{se:Evaluation}
\noindent \textbf{Evaluation Protocol:} We conduct experiments on two public PET/CT lesion segmentation datasets: AutoPET~\cite{gatidis2023autopet}, a whole-body dataset with a high metastases count and HECKTOR~\cite{oreiller2022head}, a dataset focusing on the head-neck region with fewer metastases. We benchmark GRASP using four complementary metrics: Dice~(DSC)~\cite{dice1945measures}, CC-Dice~(CC-DSC)~\cite{jaus2025every}, FP-Volume~(FPV), and FN-Volume~(FNV), computed per patient and averaged. For an easier comparison across the configurations, we compute the average rank across the four evaluation metrics for a final configuration rank. 
We experiment with three different segmentation models representing distinct architectural paradigms: 3D-UNet~\cite{cciccek20163d}, the standard encoder-decoder baseline; SegResNet~\cite{myronenko20183d}, a residual learning-based architecture; and MedNeXt-S~\cite{roy2023mednext}, a modern and efficient ConvNeXt~\cite{liu2022convnet}-inspired design, based on our obtained pretrained models.

\noindent \textbf{Implementation Details:} Training uses AdamW~\cite{loshchilov2017decoupled} (lr=$1\text{e}{-4}$, cosine annealing~\cite{loshchilov2016sgdr}), batch size 4, 300 epochs, DiceCE loss, five-fold 70:30 splits. Patch sizes: (96,96,96) for AutoPET and (64,64,64) for HECKTOR. We use 2:1 positive-to-negative sampling, including healthy samples, unlike previous works~\cite{jaus2024anatomy,kalisch2024autopet}. Our experiments are conducted on 4 NVIDIA H100 GPUs (80GB) with DDP. During inference, the feature fusion mechanism is omitted, relying solely on its regularizing effects on the decoder established during training. This approach facilitates easier model deployment and ensures that the feature fusion does not impose any additional computational burden during inference.
 
\noindent \textbf{Quantitative Results:}
We report the quantitative results in~\Cref{tab:model_results}, comparing baseline architectures against anatomical knowledge injection via a third input channel (ANA in.) and two GRASP variants (\textit{Mirror} and \textit{Mixture}). GRASP demonstrates strong performance, ranking first or second in nearly all configurations with only one exception. In a supplementary ablation on the 3D-UNet using the AutoPET dataset, we evaluated GRASP without providing anatomical pseudo-labels as additional input channels, keeping only the feature fusion block. This variant improved Dice by +1.6\% over the 2c baseline, but was still below the full 3-channel version, highlighting the complementary role of anatomical pseudo-label guidance and feature fusion.
The framework shows strong adaptability across different anatomical segmentation architectures and consistently achieves the lowest FPV on AutoPET, effectively distinguishing metabolically active tissue from actual tumors.
Performance on HECKTOR is also strong for the 3D U-Net and the SegResNet backbones. However, for MedNeXt-S, the simpler ANA in. approach slightly outperforms GRASP, suggesting that GRASP's more advanced fusion may be unnecessary for this particular backbone on this dataset.
Overall, our results show that GRASP robustly improves segmentation across diverse medical imaging tasks by integrating anatomical knowledge.

\noindent \textbf{Qualitative Results and Insights:}
We show qualitative results in~\Cref{fig:qua_3}~(left) by exploring ground-truth lesions in purple against predictions of the established model configurations in red. We display two case studies representing difficult cases due to complex topology with many small lesions (Case A) and a complex surface structure (Case B). 
We further analyze the cosine similarity of pathology features before and after fusion. 
At epoch 50, when fusion begins, similarity drops sharply, indicating that anatomical feature inclusion rapidly shifts the pathology features.
Both blocks show stabilization toward training's end, with pathology features maintaining 70-75\% similarity before and after fusion, representing a 25-30\% change due to anatomical feature inclusion. 
\input{exp/benchmark_results_alt_1}
\input{Latex_source/Latex_figure/fig_qualitative}

%% file: exp/benchmark_results_alt_1.tex
\begin{table*}
\caption{Benchmark segmentation results across two datasets. \textbf{Bold} indicates the best validation performance in each metric, while \underline{underline} denotes the second-best.}
\label{tab:model_results}
\centering
\scriptsize
\begin{tabular}{p{1.5cm}lccccc}
\toprule
\textbf{Backbones} & \textbf{Configurations} & \textbf{DSC~($\uparrow$)} & \textbf{CC-DSC~($\uparrow$)} & \textbf{FPV~($\downarrow$)} & \textbf{FNV~($\downarrow$)} & \textbf{Rank~($\downarrow$)}\\
\midrule
\multicolumn{7}{c}{\textbf{AutoPET: High Metastases Count}} \\
\midrule
\multirow{4}{*}{3D-UNet} 
& Baseline (2c) & 49.3 $\pm$ 2.9 & 31.4 $\pm$ 2.3 & \textbf{2.49 $\pm$ 2.91} & 29.96 $\pm$ 10.67 & 3 \\
& ANA in. (3c) & 52.6 $\pm$ 2.5 & 32.6 $\pm$ 2.6 & 3.16 $\pm$ 3.00 & 23.77 $\pm$ 5.52  & 3\\
& GRASP (Mixture) & \underline{53.6 $\pm$ 2.2} & \underline{33.3 $\pm$ 0.9} & 2.80 $\pm$ 2.68 & \underline{22.73 $\pm$ 6.09} & 2 \\
& GRASP (Mirror) & \textbf{54.5 $\pm$ 3.0} & \textbf{34.7 $\pm$ 2.3} & \underline{2.77 $\pm$ 2.80} & \textbf{19.75 $\pm$ 3.57} & 1 \\
\hline

\multirow{4}{*}{MedNeXt-S} 
& Baseline (2c) & 50.3 $\pm$ 2.9 & 32.1 $\pm$ 2.1 & 3.95 $\pm$ 3.55 & 36.37 $\pm$ 13.42 & 4\\
& ANA in. (3c) & 53.6 $\pm$ 3.3 & \textbf{34.6 $\pm$ 3.0} & 3.51 $\pm$ 3.23 & \underline{27.99 $\pm$ 7.87}  & 2 \\
& GRASP (Mixture) & \textbf{53.8 $\pm$ 2.9} & \underline{34.1 $\pm$ 2.4} & \textbf{2.91 $\pm$ 2.56} & 28.54 $\pm$ 9.42 & 1\\
& GRASP (Mirror) & \underline{53.6 $\pm$ 3.2} & $33.7 \pm 4.4$ & \underline{3.43$\pm$ 3.04} & \textbf{27.58$\pm$ 12.40} &2\\
\hline
\multirow{3}{*}{SegResNet} 
& Baseline (2c) & 54.1 $\pm$ 3.6 & 36.4 $\pm$ 3.5 & \underline{4.08 $\pm$ 3.23} & 32.51 $\pm$ 13.74 & 3 \\
& ANA in. (3c) & \underline{57.4 $\pm$ 3.6} & \underline{37.2 $\pm$ 3.3} & \textbf{3.39 $\pm$ 1.71} & \underline{21.89 $\pm$ 7.64}  & 2 \\
& GRASP (Mirror) &  \textbf{58.9 $\pm$ 1.2} & \textbf{39.4} $\pm$ 2.3 & 5.87 $\pm$ 4.52 & \textbf{17.08 $\pm$ 4.50} & 1\\
\midrule

\multicolumn{7}{c}{\textbf{HECKTOR: Low Metastases Count}} \\
\midrule
\multirow{4}{*}{3D-UNet} 
& Baseline (2c) & 39.9 $\pm$ 4.7 & 39.8 $\pm$ 4.7 & \textbf{0.05 $\pm$ 0.03} & 1.32 $\pm$ 0.26  & 4\\
& ANA in. (3c) & 44.8 $\pm$ 3.9 & 44.6 $\pm$ 3.9 & \underline{0.08 $\pm$ 0.04} & \underline{1.06 $\pm$ 0.15} & 2\\ 
& GRASP (Mixture) & \textbf{47.1 $\pm$ 4.0} & \textbf{46.9 $\pm$ 4.0} & 0.14 $\pm$ 0.11 & \textbf{0.75 $\pm$ 0.15} & 1 \\
& GRASP (Mirror) & \underline{45.5 $\pm$ 5.5} & \underline{45.3 $\pm$ 5.5} & 0.12 $\pm$ 0.06 & 1.16 $\pm$ 0.26  & 2\\

\hline

\multirow{4}{*}{MedNeXt-S} 
& Baseline (2c) & 53.3 $\pm$ 3.2 & 53.2 $\pm$ 3.1 & \textbf{0.27 $\pm$ 0.18} & 0.55 $\pm$ 0.16 & 3\\
& ANA in. (3c) & \textbf{56.2 $\pm$ 3.3} & \textbf{56.0 $\pm$ 3.2} & 0.39 $\pm$ 0.25 & \textbf{0.44 $\pm$ 0.13}  & 1\\
& GRASP (Mixture) & 54.3 $\pm$ 3.5 & 54.1 $\pm$ 3.4 & \underline{0.34 $\pm$ 0.17} & 0.59 $\pm$ 0.14 & 3 \\
& GRASP (Mirror) & \underline{55.2 $\pm$ 3.4} & \underline{55.1 $\pm$ 3.4} & 0.38 $\pm$ 0.15 & \underline{0.53 $\pm$ 0.20} & 2\\
\hline
\multirow{3}{*}{SegResNet} 
& Baseline (2c) & 60.5 $\pm$ 2.9 & 60.4 $\pm$ 2.8 & \textbf{0.24 $\pm$ 0.17} & 0.48 $\pm$ 0.10  & 3\\
& ANA in. (3c) & \underline{61.9 $\pm$ 4.0} & \underline{61.8 $\pm$ 4.0} & 0.68 $\pm$ 0.37 & \underline{0.45 $\pm$ 0.11}  & 2 \\
& GRASP (Mirror) & \textbf{63.3 $\pm$ 3.0} & \textbf{63.2 $\pm$ 3.0} & \underline{0.43 $\pm$ 0.27} & \textbf{0.35 $\pm$ 0.11}  & 1 \\
\bottomrule
\end{tabular}
\end{table*}

%% file: Latex_source/Latex_figure/fig_qualitative.tex
\begin{figure}[!h]
\centering
\begin{minipage}{0.66\textwidth}
  \includegraphics[width=\linewidth]{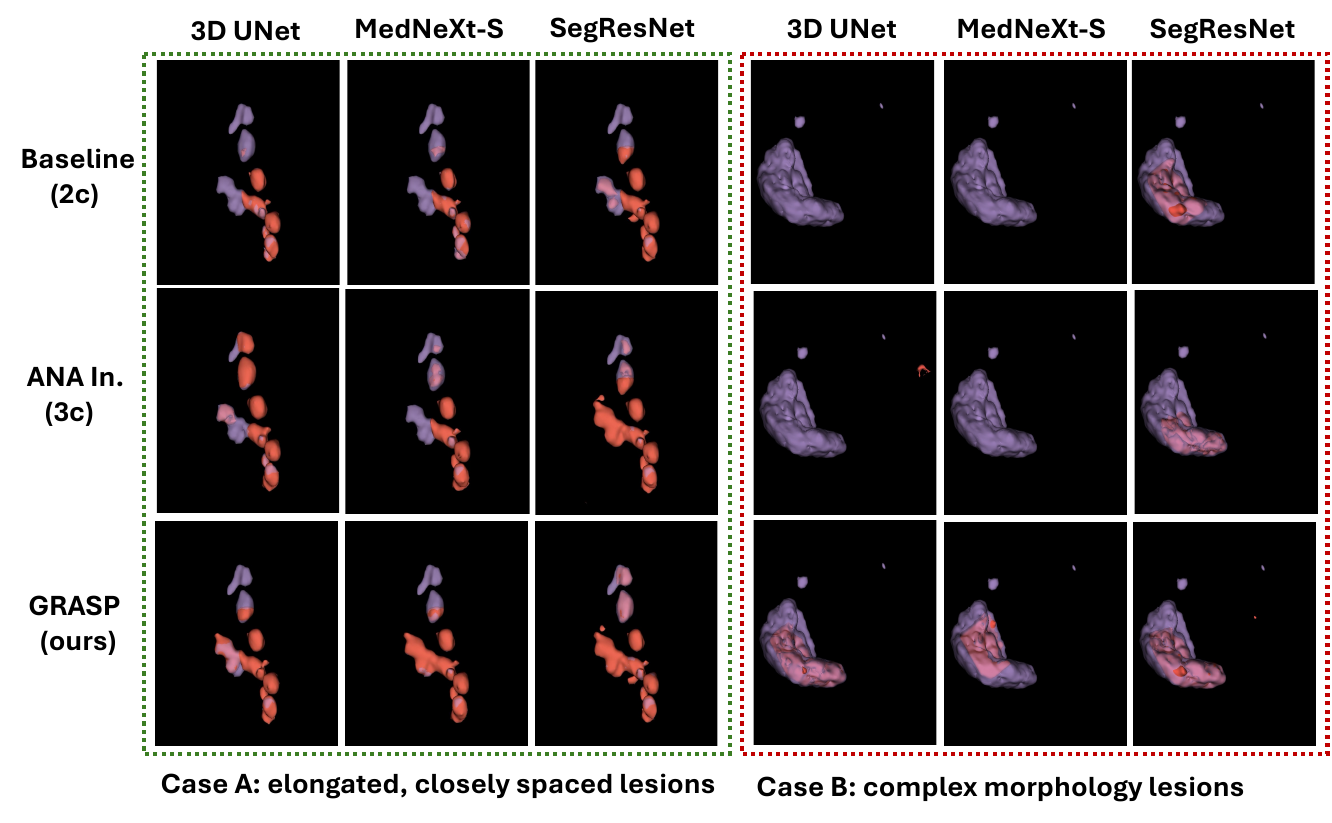}
\end{minipage}
\hfill
\begin{minipage}{0.31\textwidth}
  \includegraphics[width=\linewidth,trim=20 20 10 20,clip]{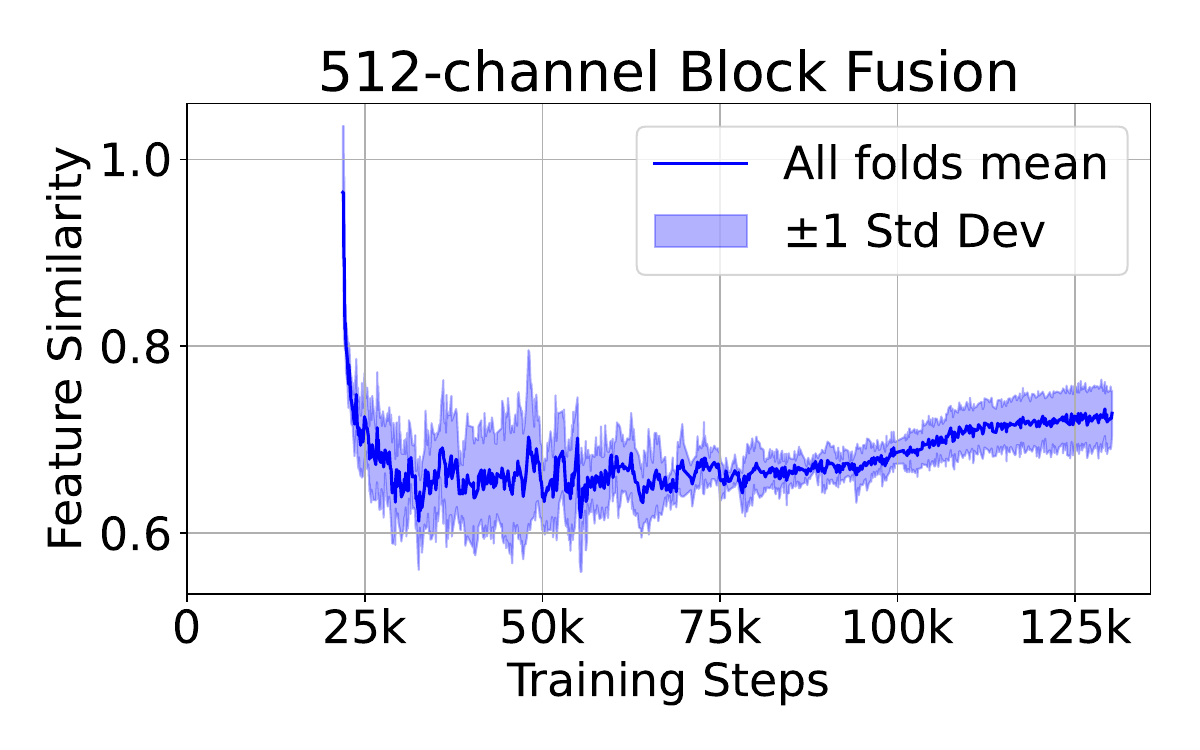}\\[1mm]
  \includegraphics[width=\linewidth,trim=20 20 10 20,clip]{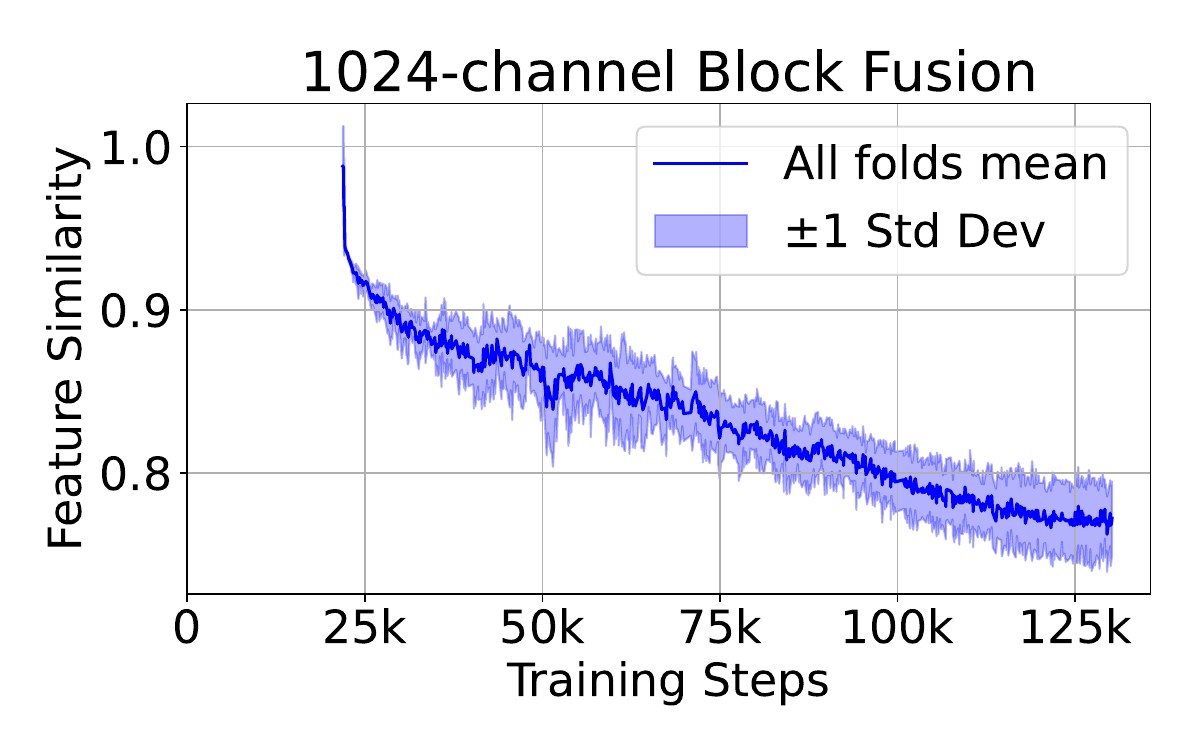}
\end{minipage}

\caption{Left: Comparison of the three backbone model configurations on two cases, with ground truth in purple and predictions in red. Right: Feature similarity of pathological features before vs. after fusion in 3D-UNet’s two fusion blocks.}
\label{fig:qua_3}
\end{figure}

%% file: Latex_source/Conclusion.tex
\section{Conclusion}
\label{se:Conclusion}
We present GRASP, a plug-and-play framework that reuses existing anatomical models by leveraging frozen anatomical encoders to enhance pathology segmentation through feature alignment. The dual injection strategy combining anatomical pseudo-labels and transformer-based bottleneck fusion consistently outperforms baseline methods across two datasets and three model architectures, achieving top rankings. GRASP's design eliminates the need for auxiliary anatomy losses, multi-stage training, or modifications of the pathology model architecture, enabling seamless integration with existing segmentation pipelines. \\

\noindent \textbf{Acknowledgements and Disclosure of Interests}:
This work was supported by the research
school “HIDSS4Health – Helmholtz Information and Data Science School for
Health”, computations were performed on the HoreKa supercomputer, funded by the Ministry of Science, Research and the Arts Baden-Württemberg and the Federal Ministry of Education and Research.
Computational resources were also provided by the Helix cluster (funded by the DFG under grant INST 35/1597-1 FUGG) and the bwUniCluster, both supported by the state of Baden-Württemberg through bwHPC.
The authors have no competing interests to declare that are relevant
to the content of this article.